\pdfoutput=1

\documentclass[11pt]{article}

\usepackage[final]{acl}

\usepackage{times}
\usepackage{latexsym}

\usepackage[T1]{fontenc}

\usepackage[utf8]{inputenc}

\usepackage{microtype}

\usepackage{inconsolata}

\usepackage{graphicx}

\usepackage{tikz}
\usetikzlibrary{matrix, arrows}
\usepackage{amsmath}
\usepackage{amsthm}
\usepackage{amssymb}
\usepackage{mathtools}
\usepackage{xspace}
\usepackage[noend]{algorithmic}
\usepackage[ruled,vlined]{algorithm2e}
\usepackage{url}
\usepackage{makeidx}
\usepackage{enumerate}
\usepackage{epstopdf}
\usepackage{booktabs}
\usepackage{color}
\usepackage{thm-restate}
\usepackage{scalerel,stackengine}
\usepackage[shortlabels]{enumitem}
\usepackage{xr}
\usepackage{fancyvrb}
\usepackage{xcolor}
\usepackage{bold-extra}
\usepackage{subfigure}
\usepackage[most]{tcolorbox}
\usepackage{fvextra}
\usepackage{float}
\usepackage{alltt}
\usepackage{soul}
\usepackage{fancyvrb}
\usepackage{multirow}
\usepackage{hyperref}
\usepackage[bottom]{footmisc}
\usepackage{bbding}
\usepackage{pifont}
\usepackage{arydshln}
\usepackage{bbm}

\newcommand\numberthis{\addtocounter{equation}{1}\tag{\theequation}}

\definecolor{title}{HTML}{064A6C}
\definecolor{boxback}{HTML}{EFEDE1}
\definecolor{burgundy}{HTML}{660033}

\newtcolorbox{promptbox}[2][]{colback=boxback, colframe=title, fonttitle=\bfseries, title=#2, boxrule=0.5mm, enhanced, flush left, halign=flush left, #1}

\newtcolorbox{samplebox}[2][]{colback=boxback, colframe=burgundy, fonttitle=\bfseries, title=#2, boxrule=0.5mm, enhanced, flush left, halign=flush left, #1}

\title{When Reasoning Meets Information Aggregation: A Case Study with \\Sports Narratives}

\author{Yebowen Hu,$^1$\thanks{Work done during Yebowen Hu's internship; Kaiqiang Song and Sangwoo Cho were full-time researchers at Tencent AI Lab, Seattle, USA at the time of this work.} Kaiqiang Song,$^2$$^*$ Sangwoo Cho,$^2$$^*$ Xiaoyang Wang,$^2$ Wenlin Yao,$^2$\\
\textbf{Hassan Foroosh,$^1$ Dong Yu,$^2$ Fei Liu$^3$}\\[0.5em]
$^1$University of Central Florida \, $^2$Tencent AI Lab, Seattle \, $^3$Emory University\\[0.3em]
$^1${\small\textsf{\{yebowen.hu, hassan.foroosh\}@ucf.edu}} $^3${\small\textsf{fei.liu@emory.edu}}\\
$^2${\small\textsf{\{riversong, swcho, shawnxywang, wenlinyao, dyu\}@global.tencent.com}}
}

\begin{document}
\maketitle

\begin{abstract}

Reasoning is most powerful when an LLM accurately aggregates relevant information. We examine the critical role of information aggregation in reasoning by requiring the LLM to analyze sports narratives. To succeed at this task, an LLM must infer points from actions, identify related entities, attribute points accurately to players and teams, and compile key statistics to draw conclusions. We conduct comprehensive experiments with real NBA basketball data and present \textsc{SportsGen}, a new method to synthesize game narratives. By synthesizing data, we can rigorously evaluate LLMs' reasoning capabilities under complex scenarios with varying narrative lengths and density of information. Our findings show that most models, including GPT-4o, often fail to accurately aggregate basketball scores due to frequent scoring patterns. Open-source models like Llama-3 further suffer from significant score hallucinations. Finally, the effectiveness of reasoning is influenced by narrative complexity, information density, and domain-specific terms, highlighting the challenges in analytical reasoning tasks.\footnote{ \url{https://github.com/YebowenHu/SportsGen}}

\end{abstract}

\section{Introduction}
\label{sec:intro}

Accurately aggregating information is essential for reasoning~\cite{sprague2024musr,wang2023hypothesis,yasunaga2024large,hu2024defineenhancingllmdecisionmaking}. This is especially true when reasoning over longitudinal data~\cite{kidziński2021modeling}. For example, when assessing patient outcomes over years to determine the effectiveness of new medications, models must accurately track changes to provide insights into trends, patterns, and potential causal relationships. While LLMs hold significant potential in reasoning, understanding how they aggregate information to support reasoning is key to ensuring their robustness in decision-making.

\begin{figure*}
\centering
\includegraphics[width=6.25in]{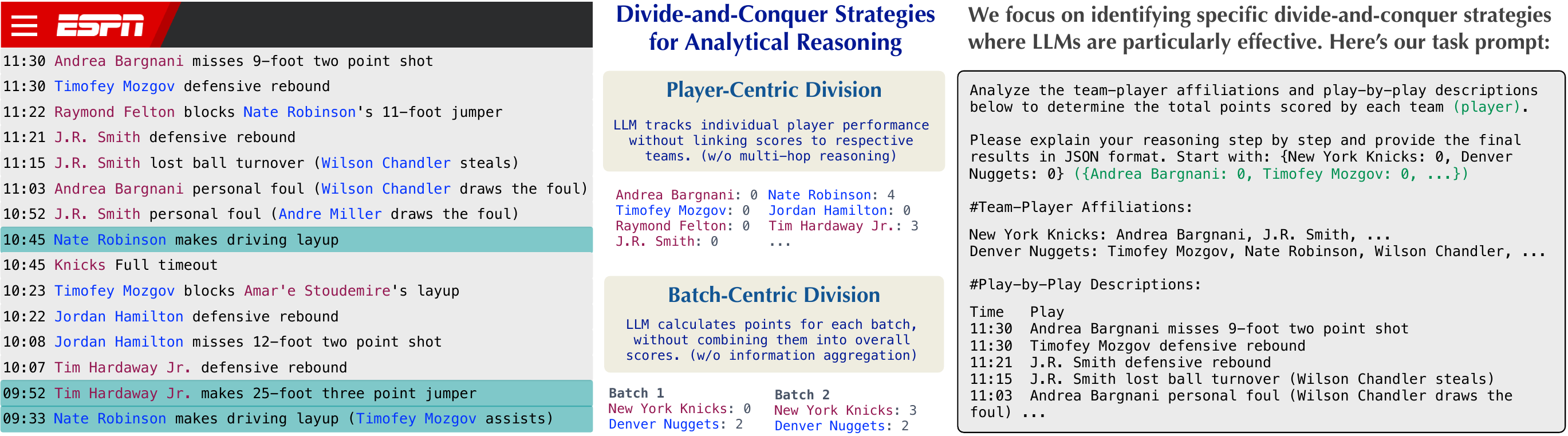}
\vspace{-0.05in}
\caption{ESPN's NBA game play-by-play descriptions. We are particularly interested in exploring whether LLMs can perform analytical reasoning in a more focused and manageable context using divide-and-conquer strategies.}
\label{fig:example}
\vspace{-0.1in}
\end{figure*}

Sports data have recently emerged as a popular testing ground for LLMs~\cite{srivastava2023imitation,wang2023tacticai,li2024sportsqa,yang2024crag}.
Unlike health data, sports narratives are readily accessible from diverse sources, e.g., ESPN, CBS Sports, BBC Sport, and Sports Illustrated.\footnote{\url{espn.com} \,\, \url{cbssports.com} \,\, \url{bbc.com/sport} \,\, \url{si.com}} These narratives provide detailed event descriptions over extended game periods, making them an excellent case study to explore how LLMs reason with longitudinal data. Figure~\ref{fig:example} presents a snippet from an NBA basketball game. The play-by-plays capture key actions of the game, such as passes, shots, and fouls, along with timestamps and team-player affiliations. \emph{Our research involves using LLMs to track scoring actions within these detailed game descriptions and to link entities for comprehensive game analysis.} These games include a mix of expected and unexpected events, presenting unique challenges for reasoning tasks.

LLMs's reasoning abilities, including multi-hop, deductive, inductive, and abductive reasoning, have garnered increasing attention~\citep{Hu:2023aa,Shi:2023aa,Zhao:2023aa,Yang:2024aa,Yang:2024ab,Li:2024aa,Duan:2024aa,Chu-Carroll:2024aa}. For example, \citet{Sprague:2023aa} introduce a dataset designed to evaluate language models on multi-step reasoning with 1,000-word murder mysteries and object placement. \citet{Yang:2024aa} explore latent multi-hop reasoning and found evidence of its use in fact composition. \citet{berglund2024reversal} study the reversal curse that LLMs trained on `A is B' fail to learn that `B is A.' Distinguishing from previous studies, our approach introduces a quantitative aspect to reasoning with game narratives, where the LLM must summarize key statistics to derive conclusions.

We focus on \emph{analytical reasoning} in this work, which involves identifying relationships between elements, drawing inferences and synthesizing information~\cite{savage2023diagnostic}. Notably, we evaluate if LLMs perform better in a more focused and manageable context when using a \emph{divide-and-conquer} strategy. Sports narratives, for instance, can be segmented along multiple dimensions. We explore division methods that \emph{simplify multi-hop reasoning} or \emph{facilitate information aggregation}, thereby easing the load on LLMs. We further introduce a new Discounted Cumulative Accuracy (DCA) metric. DCA differs from traditional accuracy by allowing a small margin of error. This helps us identify scenarios where LLMs significantly hallucinate, rather than grounding predictions on the narratives. Our experiments reveal multiple limitations of state-of-the-art LLMs in using divide-and-conquer strategies for analytical reasoning.

With data synthesis, we further stress test LLMs' reasoning abilities in complex and novel scenarios. LLMs are required to handle a range of narratives that differ in length, complexity, and information density, and may also involve a shift from natural language to symbolic reasoning. Our approach, \textsc{SportsGen}, enhances the control over narrative complexity when compared to human-written game narratives and few-shot prompting. It contributes to ongoing research in data synthesis~\cite{veselovsky2023generating,arora2023language,kang2024humanintheloop,divekar2024synthesizrr} by creating synthetic sports narratives that help evaluate the analytical capabilities of LLMs. Our research contributions are summarized as follows.

\begin{itemize}[topsep=3pt,itemsep=-1pt,leftmargin=*]
     
\item We investigate LLMs' reasoning abilities by analyzing play-by-play sports narratives. This involves tracking scoring actions, inferring points from actions, recognizing players and teams, and summarizing key statistics to draw conclusions. By exploring how LLMs aggregate information, we gain insights into their potential in longitudinal studies, such as patient health management, where the LLM must identify and interpret recurring patterns to make critical decisions.

\item We observe that LLMs struggle more with accurately aggregating points than assigning points to players and teams. Basketball's frequent scoring poses a challenge for tracking actions. Moreover, when input narratives are short and instructions are lengthy, LLMs may overlook the narrative and hallucinate game points. Our \textsc{SportsGen} approach has significant implications for simulating complex reasoning scenarios that involves synthesizing both text and numerical data.

\end{itemize}

\section{Related Work}
\label{sec:related}

Success in reasoning hinges on effective content selection. Recent studies have placed a growing emphasis on LLM's reasoning capabilities, exploring various types such as deductive, inductive, abductive, and multi-hop reasoning~\citep{Bostrom:2021aa,Huang:2022aa,Hu:2023aa,Zhao:2023aa,Holliday:2024aa}. Efforts to enhance LLMs' reasoning abilities include prompting, supervised fine-tuning, and adjustments to decoding~\citep{wei2023chainofthought,chen2023skillsincontext,Zhao:2023aa,Burnell:2023aa,zheng2023step,Cheng:2024aa,Ahn:2024aa,Wang:2024aa}. In our paper, we propose a quantitative approach to reasoning by requiring the LLM to identify and consolidate relevant facts. Similar to how frequency indicates salient content in multi-document summarization~\cite{fabbri-etal-2019-multi,mao-etal-2020-multi,gholipour-ghalandari-etal-2020-large,lebanoff-etal-2021-modeling, THOMSON2023101482}, we conjecture that how LLMs aggregate information is critical in addressing complex reasoning challenges.

Numerical reasoning is applied in financial question answering and solving mathematical word problems, and numerous datasets have been developed for these purposes~\cite{amini2019mathqa,cobbe2021training,patel2021nlp,zhu2021tatqa,chen2022finqa,liu2023improving,zhao2023docmatheval,lu2023dynamic}. Our study explores a new angle—investigating LLMs' ability to analytically solve problems using divide-and-conquer strategies~\cite{lee-kim-2023-recursion,you-etal-2023-idealgpt}. Sports data are complex and multifaceted. Our focus is on pinpointing specific divide-and-conquer strategies where LLMs are particularly effective.

Sports data plays a pivotal role in various language tasks, such as real-time summarization of games, data-to-text generation, and analysis of commentator bias~\cite{wiseman-etal-2017-challenges,edouard-etal-2017-youll,van-der-lee-etal-2017-pass,puduppully-etal-2019-data,merullo-etal-2019-investigating,huang-etal-2020-generating,hu2024large,hu2024sportsmetrics}. It also influences the analysis of domains like game reviews and gameplay logs, enhancing our understanding of gameplay commentary dynamics~\cite{gamesandnlp-2020-games,kicikoglu-etal-2020-aggregation,gu-etal-2022-revisiting,furman-etal-2022-sequence}. Building on research by \citet{hu2024sportsmetrics}, our study explores how well LLMs can track scoring actions across lengthy narratives and accurately link entities for in-depth game analysis. With \textsc{SportsGen} for data synthesis~\cite{veselovsky2023generating,arora2023language,divekar2024synthesizrr}, we take a step further by creating diverse narratives that vary in style, complexity, and level of detail to better assess LLM effectiveness. 


\section{Analytical Reasoning with Divide-and-Conquer Strategies}
\label{sec:divide}

Our research uses analytical reasoning to accurately calculate team points from sports narratives. We specifically focus on ESPN's NBA game play-by-play descriptions (Figure~\ref{fig:example}). On average, an NBA game consists of 466 plays and 6,229 tokens, with the longest narrative containing up to 7,322 tokens. 
To succeed at this task, an LLM must understand the basic rules of the game, distinguishing between scoring plays, such as three-pointers and field goals, and non-scoring actions like passing. Further, the model must correctly attribute each scoring action to the appropriate player and team, which requires nuanced reasoning. Finally, the LLM must aggregate all the scored points to determine the final scores for the teams. This process tests the LLM in three critical areas: knowledge acquisition, point referencing, and effective score aggregation. Errors in any of these areas can lead to incorrect results.

\begin{figure*}
\centering
\includegraphics[width=6.3in]{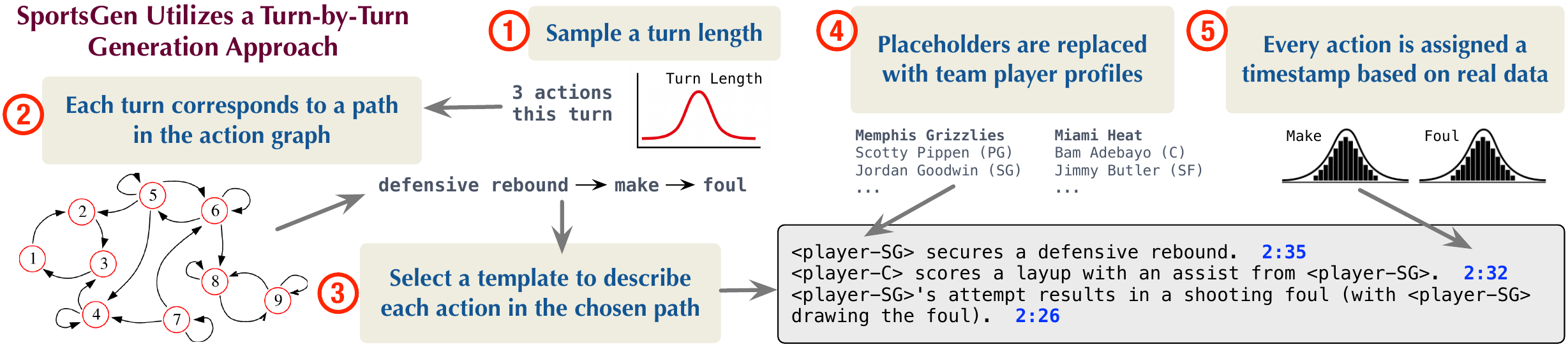}
\vspace{-0.2in}
\caption{\textsc{SportsGen}, a new method that synthesizes sports narratives by modeling game dynamics.}
\label{fig:generation}
\end{figure*}

\vspace{-0.05in}\paragraph{Player-Centric Division.} We are particularly interested in exploring whether LLMs can perform analytical reasoning in a more focused and manageable context using a divide-and-conquer strategy. For instance, sports narratives can be segmented along multiple dimensions. In our \emph{\textbf{player-centric division}}, the LLM assesses player performance based on play-by-play descriptions. Here, the LLM receives a list of players, without their team affiliations, and is tasked with tracking and summarizing each player's performance throughout the game. The final results are presented in JSON format. Following this, we link the individual players' scores with their respective teams to calculate the overall team scores. This player-centric approach employs LLMs to track individual player performances, allowing for a direct calculation of team scores from player data. It lessens the burden on LLMs to use multi-hop reasoning for inferring team scores from player scores.

\vspace{-0.05in}\paragraph{Batch-Centric Division.} Instead of processing the play-by-play narrative at once, we break it down into smaller, non-overlapping batches, each containing $K$ plays. In our \emph{\textbf{batch-centric division}} approach, each batch is analyzed independently by the LLM to determine the team points for that segment. This approach allows the LLM to narrow its focus, helping it easily identify where errors might be occurring. It also eases the burden on LLMs, as they are not required to aggregate points to determine total scores from batch scores; instead, this is handled through direct calculation. We compare this approach to \emph{\textbf{monolithic processing}}, where an LLM processes the entire game narrative in one attempt. Interestingly, one would expect that when LLMs are provided with data in smaller batches, they would perform better in this more granular and manageable setting. In \S\ref{sec:experiments}, we report notable findings that highlight the limitations of state-of-the-art LLMs when using these divide-and-conquer strategies.

We focus on identifying specific data divisions where LLMs are particularly effective. Real-world data can be complex and multifaceted. By focusing on sports narratives, we aim to understand the limitations of LLMs in reasoning and information aggregation, since complex reasoning often require LLMs to aggregate relevant information effectively. Our research does not introduce new methods of reasoning. Instead, it uses chain-of-thought prompting, which enables Transformers to tackle inherently serial problems~\cite{li2024chain}. In the following section, we propose synthesizing sports narratives, which will allow us to challenge LLMs in varying and unseen scenarios.

\section{Creating Data with \textsc{SportsGen}}
\label{sec:synthesis}

We introduce \textsc{SportsGen}, a new method that synthesizes sports narratives by modeling game dynamics. \textsc{SportsGen} offers a platform to assess LLMs' reasoning capabilities in novel scenarios that may not exist in real data. Since synthesized narratives have not been encountered by LLMs during pretraining, they also serve as a valuable benchmark for future LLM assessments. Previously, data synthesis has been utilized in fine-tuning, distilling, and evaluating LLMs for tasks such as hallucination detection, topic classification, sarcasm and humor detection~\cite{veselovsky2023generating,tang2023does,arora2023language,kang2024humanintheloop,zhang2024halugen,divekar2024synthesizrr}. Our work expands on these efforts by focusing on the specific challenges in generating game narratives.\footnote{Spatial information, such as player positions on the field, is not included in our synthesized narratives, as it falls outside our current scope.}

\vspace{-0.05in}\paragraph{Setting Up Team Profiles.} We create profiles for two basketball teams, Team $\mathcal{T}_1$ and Team $\mathcal{T}_2$, for a specific game. Each team has five players, filling the roles of point guard (PG), shooting guard (SG), small forward (SF), power forward (PF), and center (C). Each team has an {efficiency score} $\mathcal{E}$ from 0 to 100, which reflects the points they score per 100 possessions.\footnote{Efficiency scores for basketball teams are obtained from ESPN. We aim to model team effectiveness in goal attempts; future research may consider individual player metrics.} We start with real NBA teams for these profiles, and later users can select either real or synthetic players to create fantasy teams.

\vspace{-0.05in}\paragraph{Turn-by-Turn Game Generation.} Our \textsc{SportsGen} renders realistic sports narratives using a turn-by-turn approach, which mirrors the back-and-forth dynamics seen in live sports games. In each turn, a team controls the ball and performs a series of plays\footnote{In our paper, `plays' and `actions' are used interchangeably to describe significant game actions.} such as passes, shots, and fouls, which are key actions during the game. After each turn, possession generally shifts to the opposing team. Each quarter consists of alternating turns between the two teams, with each turn being given an allotted time frame. For example, a turn might unfold as follows: ``Lakers take possession, make a quick pass, and take a shot that misses,'' all within about 15 seconds. Each basketball game lasts 48 minutes. Our approach continues to add turns until nearing the end of a quarter's time limit. If the ending turn exceeds this limit, it will be truncated.

\vspace{-0.05in}\paragraph{Building an Action Graph.} When a team has the ball, they are on offense and use actions such as passing, dribbling, shooting to score points. The defense tries to stop them by blocking, stealing, and rebounding. We refer to each sequence of these actions as a \textbf{turn}, and characterize it using a Markov graph. For example, a turn might start with a \emph{defensive rebound}, followed by a \emph{missed shot}, and then conclude. In the graph, each significant action is represented as a node, and transitions between nodes show how team members cooperate and execute their tactics. The graph begins and ends with special nodes that mark the start and end of a turn. Our model includes 16 key actions; an example is provided in Appendix \ref{fig:action_graph}. This action graph helps to visualize the game's flow within each turn.

\vspace{-0.05in}\paragraph{Action Templates.} Each action is linked to specific templates that describe it. E.g., `\emph{<player-PF> misses a 20-foot jumper.}' and `\emph{<player-PF> fails to convert a 14-foot pull-up jump shot.}' correspond to the action `miss'. To gather these templates, we first search NBA data using keywords selected by experts to identify plays that describe each action. We then provide these initial templates to GPT-4o to increase their lexical diversity and enhance the narrative quality. An example of actions and their templates is shown in Table~\ref{tab:templates}. During generation, each template linked to a sampled action has an equal probability of being used. The transitions between actions are determined by analyzing real game narratives to calculate maximum likelihood estimates.

\vspace{-0.05in}\paragraph{Controlling Narrative Complexity.} We adjust the complexity of the game by changing the number of actions per turn. Longer turns can extend the overall narrative, whereas shorter turns create more frequent scoring opportunities, as each turn often results in an attempt to score. In NBA narratives, a turn might have anywhere from 1 to 10 actions, with an average of 1.65 actions. This results in a scoring to non-scoring action ratio $\mathcal{R}$ = 1:3. To model turn length, we employ a Gaussian distribution $\mathcal{G}$ parameterized by the mean ($\mu$) and standard deviation ($\sigma$), which we calibrate using NBA data. During synthesis, we vary $\mu$ to generate narratives with different turn lengths while keeping $\sigma$ stable. This approach allows us to produce narratives with realistic scoring ratios such as $\mathcal{R}$ = 1:2, 1:3, 1:4, or 1:5, effectively capturing the dynamics of actual basketball games.

\begin{table}
\setlength{\tabcolsep}{1pt}
\renewcommand{\arraystretch}{0.95}
\centering
\begin{scriptsize}
\begin{tabular}{l}
\toprule
\multicolumn{1}{c}{\texttt{\textbf{Action - ``make''}}}\\
\textbullet\, \texttt{<player-SG> successfully makes a 2-foot two point shot with}\\
\,\,\, \texttt{the assistance of <player-SF>.}\\
\textbullet\, \texttt{<player-SF> executes a delicate finger roll layup.}\\
\textbullet\, \texttt{<player-PG> hit a 26-foot three pointer.}\\
\textbullet\, \texttt{<player-PG> executes a 21-foot step back jumpshot.}\\
\textbullet\, \texttt{<player-SG> sinks a 26-foot three-point shot with an assist}\\
\,\,\, \texttt{from <player-C>.}\\
\textbullet\, \texttt{<player-SF> executes a tip shot...}\\
\midrule
\multicolumn{1}{c}{\texttt{\textbf{Action - ``miss''}}}\\
\textbullet\, \texttt{<player-PG> fails to make a 25-foot standard jump shot.}\\
\textbullet\, \texttt{<player-C> fails to convert a 20-foot jumper.}\\
\textbullet\, \texttt{<player-SG> unsuccessfully attempts a three point jumper.}\\
\textbullet\, \texttt{<player-PF> misses a 20-foot jumper.}\\
\textbullet\, \texttt{<player-PG> misses a 4-foot driving layup...}\\
\midrule
\multicolumn{1}{c}{\texttt{\textbf{Action - ``block''}}}\\
\textbullet\, \texttt{<player-PG> individual defensive play resulted in a personal}\\
\,\,\, \texttt{block (<player-SG> draws the foul).}\\
\textbullet\, \texttt{<player-PF> executes a personal block as <player-PG> draws}\\
\,\,\, \texttt{the foul.}\\
\textbullet\, \texttt{<player-PG> successfully executes a personal block against}\\
\,\,\, \texttt{the opponent, drawing the foul in the process.}\\
\textbullet\, \texttt{<player-PG> commits a shooting block foul and <player-G> is}\\
\,\,\, \texttt{fouled during the play...}\\
\bottomrule
\end{tabular}
\end{scriptsize}
\vspace{-0.05in}
\caption{Each action is linked to specific templates that describe it. The templates are enhanced by GPT-4o to increase their lexical diversity and narrative quality. 
}
\label{tab:templates}
\end{table}

\vspace{-0.05in}\paragraph{Rendering the Narrative.} We begin by selecting turn lengths from the Gaussian function $\mathcal{G}$ (Figure~\ref{fig:generation}). Each turn is then rendered by tracing a path through the action graph that matches the designated length. We evaluate these paths for quality and reject any that are unlikely in a basketball game according to a set of expert rules. The probability of a turn leading to a score is determined using the Efficiency Score ($\mathcal{E}$) for the involved team. From this, we sample a `1' to indicate scoring and `0' otherwise. A scoring turn must include a `make' action to indicate a successful score, and we record the points earned by the team in the box score. 

We repeatedly sample from the action graph to find a path that meets our length and scoring criteria. We then select a template to describe each action in the chosen path. Next, placeholders are replaced with relevant team player profiles, e.g., `player-PG' is replaced with the actual point guard's name. Each action is tagged with a timestamp indicating the time elapsed since the previous action. We calculate these time intervals using real data and create a Gaussian $\mathcal{G}_a$ for each action type. We alternate between the profiles of the two teams involved in the game. This process is repeated until we have created enough turns to complete one game quarter, after which we proceed to generate the remaining quarters to complete the full game narrative.

\begin{table*}
\setlength{\tabcolsep}{7pt}
\renewcommand{\arraystretch}{1}
\centering
\begin{small}
\begin{tabular}{llrrrrrrrrr}
& & \multicolumn{6}{c}{\textbf{Human-Written Sports Narratives}} & \multicolumn{3}{c}{\textbf{Synthetic Narratives}}\\
\cmidrule(lr){3-8}\cmidrule(lr){9-11}
& \textbf{Model} & \texttt{\textbf{DnC-1}} & \texttt{\textbf{DnC-3}} & \texttt{\textbf{DnC-10}} & \texttt{\textbf{DnC-30}} & \texttt{\textbf{DnC-P}} & \texttt{\textbf{Mono}} & \texttt{\textbf{DnC-10}} & \texttt{\textbf{DnC-P}} & \texttt{\textbf{Mono}} \\
\midrule
\multirow{6}{*}{\rotatebox[origin=c]{90}{\textbf{Accuracy}}} & \texttt{Llama3-8B-Inst} & 7.92 & 33.79 & 33.42 & {38.61} & 4.70 & 4.43 & 11.67 & 5.86 & 3.78 \\
& \texttt{Gemini-Pro-1.5} & 48.27 & {67.20} & 64.36 & 32.43 & 8.79 & 4.58 & 15.00 & 11.67 & 3.75 \\
& \texttt{GPT-3.5-Turbo} & 14.60 & {56.68} & 44.43 & 21.41 & 6.68 & 2.00 & 5.00 & 8.33 & 1.25\\
& \texttt{Llama3-70B-Inst} & 22.40 & 63.12 & {80.45} & 84.53 & 21.29 & 17.45 & \textbf{35.00} & 24.17 & 9.17\\
& \texttt{GPT-4o} & \textbf{72.77} & \textbf{84.28} & \textbf{88.61} & \textbf{84.90} & \textbf{68.19} & 45.54 & 28.33 & \textbf{52.08} & 37.92\\
& \texttt{Claude-3-Opus} & 60.64 & 75.99 & {84.16} & 81.31 & 61.39 & \textbf{67.20} & 31.67 & 42.50 & \textbf{54.58}\\
\midrule
\multirow{6}{*}{\rotatebox[origin=c]{90}{\textbf{DCA}}} & \texttt{Llama3-8B-Inst} & 39.18 & {87.96} & 86.18 & 86.79 & 36.11 & 41.5 & 71.21 & 55.94 & 38.08 \\
& \texttt{Gemini-Pro-1.5} & 91.58 & {95.83} & 94.94 & 81.75 & 46.21 & 18.56 & 81.52 & 69.09 & 24.58\\
& \texttt{GPT-3.5-Turbo} & 70.66 & {93.27} & 89.60 & 74.31 & 55.64 & 29.00 & 62.27 & 60.80 & 25.19\\
& \texttt{Llama3-70B-Inst} & 55.54 & 92.20 & \textbf{96.12} & 97.52 & 63.81 & 74.18 & 87.73 & 77.01 & 68.90\\
& \texttt{GPT-4o} & 93.06 & \textbf{97.52} & \textbf{98.41} & \textbf{97.66} & \textbf{92.38} & 86.70 & 82.42 & 78.79 & 82.69\\
& \texttt{Claude-3-Opus} & \textbf{94.76} & 97.07 & {98.28} & 97.60 & 91.39 & \textbf{93.56} & \textbf{90.00} & \textbf{85.00} & \textbf{89.32}\\
\bottomrule
\end{tabular}
\end{small}
\vspace{-0.05in}
\caption{Results of top LLMs in calculating team points from sports narratives, divided by batches of 1, 3, 10, or 30 plays (DnC-\{1,3,10,30\}), by individual players (DnC-P), and monolithic processing (Mono).
}
\label{tab:results_real}
\vspace{-0.1in}
\end{table*}

\section{Experiments}
\label{sec:experiments}

Sports data are accessible from a variety of sources such as ESPN, BBC Sport, CBS Sports, and Sports Illustrated. This work utilizes the basketball play-by-plays provided by \citet{hu2024sportsmetrics} consisting of 28,492 NBA games from 2002 to 2023, gathered from ESPN. Due to resource constraints, we subsampled the original dataset to $D_\mathcal{H}$, corresponding to the number of human-written narrative quarters. Further, our proposed method, \textsc{SportsGen}, generates synthesized game narratives with varied  scoring to non-scoring (S:NS) ratios including 1:2, 1:3, 1:4, and 1:5. $D_{\mathcal{S}}$ represents the number of game quarters synthesized by \textsc{SportsGen}.

\vspace{-0.05in}\paragraph{Accuracy.} Our goal is to enable LLMs' to track and predict the total points scored by each team at the end of a game quarter. We choose quarter-level evaluation because current LLMs do not perform well at tracking points for full games. 
The \emph{\textbf{accuracy}} metric strictly evaluates whether the model's predictions matches with the actual scores for each quarter and team. Here, each team is considered as a separate data point. For human-written narratives, our dataset includes detailed box scores sourced from ESPN. For our \textsc{SportsGen} generated narratives, ground-truth scores are obtained during the generation process.

\vspace{-0.05in}\paragraph{Discounted Cumulative Accuracy (DCA).} 
``Exact match'' accuracy is not an ideal metric for tracking numerical values, as it is strict and can unfairly penalize LLMs. To address this, we introduce the \emph{\textbf{discounted cumulative accuracy}} metric, which differs from traditional accuracy by allowing a small margin of error. DCA rewards predictions that are close to the true value, with diminishing gains as predictions deviate further. This approach draws inspiration from the DCG~\cite{10.1145/582415.582418}, a highly effective metric in information retrieval. It provides a balanced evaluation by cumulatively adjusting rewards, allowing for a fairer assessment of a system's performance.

Concretely, we denote the tolerance level as $\texttt{T}$ (Eq.(\ref{eq:dca})). At zero tolerance ($\texttt{T}$ = 0), the DCA metric becomes standard accuracy. As the tolerance level increases, performance differences between models become less pronounced, making this a more forgiving evaluation metric. During each time step $t$, we apply a discounting factor of $1 - \frac{t}{\texttt{T}}$. With a tolerance of 10 points ($\texttt{T}=10$), this factor decreases gradually from 1 down to 0 in steps of 0.1. 

We use $p_t$ to denote the proportion of instances where the prediction error is exactly $t$ points from the true value (Eq.(\ref{eq:p_t})). Let $N$ represent the total number of instances evaluated. For each instance, $\lvert s_n - s_n^*\rvert$ measures the absolute difference between the predicted and actual values. If this error surpasses $\texttt{T}$, the instance does not contribute to the score. Higher accuracy and DCA scores indicate better-performing systems. 
\begin{align*}
\textstyle
\texttt{DCA} &= \sum_{t=0}^\texttt{T} p_t(1-\frac{t}{\texttt{T}}) 
\numberthis\label{eq:dca}\\
p_t &= \frac{1}{N}\sum_{n=1}^N \mathbbm{1}_{\{\lvert s_n - s_n^* \rvert = t\}} \numberthis\label{eq:p_t}
\end{align*} 
In the sections below, we evaluate leading LLMs on their analytical reasoning with game narratives using various divide-and-conquer strategies. We also compare human-written sports narratives against our \textsc{SportsGen} narratives to assess LLM effectiveness.

\subsection{Divide-and-Conquer Results}
\label{sec:divide-and-conquer}

In Table~\ref{tab:results_real}, we compare the analytical reasoning capabilities of leading LLMs.\footnote{The lineup includes both proprietary and open-source models: Llama3-8B-Instruct, Llama3-70B-Instruct ~\citep{llama3}, OpenAI's GPT-4o and GPT-3.5-Turbo~\citep{gpt4o}, Gemini-Pro-1.5~\citep{geminiteam2023gemini}, and Anthropic's Claude-3-Opus~\citep{claude-3}. These LLMs are available to the research community from March to May 2024. The models can perform complex analyses, handle multi-step procedures, and tackle higher-order math and reasoning tasks.}
Our evaluation shows that both GPT-4o and Claude-3-Opus are effective for this task, with Claude-3-Opus outperforming especially in monolithic processing `Mono' scenarios, where the entire game narrative is processed as a whole. Here, Claude-3-Opus achieves 67.20\% accuracy and a DCA score of 93.56\%. This indicates that the model provides exactly accurate results in over 60\% of cases and is within a minimal error margin ($\texttt{T}$ = 0) in >90\% of the scenarios. This superior performance may stem from Claude-3-Opus's advanced logical reasoning and computational abilities. Of the three variants in the Claude-3 model family (Haiku, Sonnet, and Opus), Opus is the most sophisticated and costly. The expense for using this API is \$15 / million tokens for input and \$75 / million tokens for output.

On the other hand, GPT-4o excels when using the divide-and-conquer (DnC) strategy on human-written sports narratives. It achieves the highest accuracy at 88.61\% and a DCA score of 98.41\% with a batch size of 10 (DnC-10). This suggests that incorporating innovative strategies like DnC can potentially narrow the performance gaps among top-performing models. The open-source Llama3-70B-Inst model performs on par with GPT-4o and Claude-3-Opus. However, lesser models such as Llama3-8B-Inst, Gemini-Pro-1.5, and GPT-3.5-Turbo lag behind, with Llama3-8B-Inst delivering the lowest accuracy and DCA scores.

\vspace{-0.05in}\paragraph{Batch-Centric vs. Player-Centric Division.} We test the performance of LLMs using different batch sizes by dividing the game's play-by-plays into batches of 1, 3, 10, or 30 plays, denoted as DnC-\{1,3,10,30\} in Table~\ref{tab:results_real}. Our findings revealed two key insights. First, the optimal batch size varies between models. High-performing models such as Claude-3-Opus, GPT-4o, and Llama3-70B-Inst show peak performance with a batch size of 10. In contrast, less robust models perform best with smaller batches, such as 3 plays per batch. This suggests that finding the right balance between accuracy per batch and the total number of batches is crucial for achieving optimal results.

Notably, using the smallest batch size (1 play) did not yield the best performance. This was primarily because play-by-play data in small batches were often overshadowed by system messages and instructions, resulting in models, particularly the Llama3-8B-Inst, generating hallucinated scores. Increasing the batch size generally led to more scoring errors for all models. This emphasizes the need for a strategic approach to selecting batch sizes as well as enhancing models' instruction following abilities to improve their robustness. The DnC-P method slightly enhances LLM accuracy compared to processing the narrative as a whole (`Mono'), although it is not as effective as segmenting the narrative into smaller batches. Our findings suggest that while LLMs excel in reasoning, their performance in information aggregation is weaker.

\begin{table}
\setlength{\tabcolsep}{3.8pt}
\renewcommand{\arraystretch}{1.05}
\centering
\begin{small}
\begin{tabular}{lrrrrr}
& \multicolumn{1}{c}{\textbf{Accu.}} & \multicolumn{4}{c}{\textbf{Disc. Cumu. Accu.}}\\
\cmidrule(lr){2-2} \cmidrule(lr){3-6}
\textbf{Model} & \multicolumn{1}{c}{\textbf{\texttt{T=0}}} & \multicolumn{1}{c}{\textbf{\texttt{T=1}}} & \multicolumn{1}{c}{\textbf{\texttt{T=3}}} & \multicolumn{1}{c}{\textbf{\texttt{T=5}}} & \multicolumn{1}{c}{\textbf{\texttt{T=10}}} \\
\midrule
\texttt{Llama3-8B-Inst} & 4.43 & 8.53 & 17.22 & 25.48 & 41.50 \\
\texttt{Gemini-Pro-1.5} & 4.58 & 6.44 & 10.15 & 12.89 & 18.56 \\
\texttt{GPT-3.5-Turbo} & 2.00 & 3.55 & 8.17 & 13.69 & 29.00 \\
\texttt{Llama3-70B-Inst} & 17.45 & 29.58 & 47.65 & 59.59 & 74.18 \\
\texttt{GPT-4o} & 45.54 & 52.72 & 68.94 & 77.27 & 86.70 \\
\texttt{Claude-3-Opus} & \textbf{67.20} & \textbf{72.22} & \textbf{83.08} & \textbf{88.41} & \textbf{93.56} \\
\bottomrule
\end{tabular}
\end{small}
\vspace{-0.05in}
\caption{We experiment with varying tolerance levels (T = 0, 1, 3, 5, 10) to assess the impact of this parameter. Adjusting the tolerance threshold generally does not alter the relative rankings of the models. 
}
\label{tab:results_DCA}
\vspace{-0.15in}
\end{table}

\vspace{-0.05in}\paragraph{Accuracy vs. DCA.} In Table~\ref{tab:results_DCA}, we experiment with varying tolerance levels (T = 0, 1, 3, 5, 10) to assess the impact of this parameter. We note that adjusting the tolerance threshold generally does not alter the relative rankings of the models. However, increasing the tolerance tends to narrow the performance gap among different models. Importantly, accuracy scores alone may not be the most suitable metric for evaluation. Lower-performing models often result in accuracy in the low single digits. For example, the model Gemini-Pro-1.5 reports an accuracy of just 4.58\%. Its performance remains the lowest even when tolerance is increased to 10 points (T=10), resulting in the lowest DCA score of 18.56\%. Such findings highlight Gemini-Pro-1.5’s considerably weaker analytical reasoning capabilities even under tolerance settings.

\subsection{Human Versus Synthetic Narratives}
\label{sec:human_synthetic}

In Table~\ref{tab:human_synthetic}, we compare human-written sports narratives with those generated by SportsGen and a few-shot prompting approach. Due to resource limits, we set $D_\mathcal{H}$=400 for human narratives and $D_\mathcal{S}$=480 for each synthetic narrative set across all experiments. Few-shot prompting, detailed in Appendix~\ref{sec:example_prompts}, uses existing play-by-plays to guide GPT-4o in creating synthesized game quarter. This method can result in repetitive scoring actions, with a high {S:NS} Ratio of 1:0.4, deviating from typical basketball games. In contrast, \textsc{SportsGen} offers enhanced controllability and conveniently generates box scores during generation. We adjust the turn length to produce games with varying information densities, featuring S:NS ratios of 1:2, 1:3, 1:4, and 1:5. With an S:NS ratio of 1:3, it mirrors the ratio found in actual human narratives and resulting in an average of 1,700 tokens per narrative. This makes \textsc{SportsGen} a more practical option for creating realistic sports narratives.

\begin{table}[t]
\setlength{\tabcolsep}{3pt}
\renewcommand{\arraystretch}{1}
\centering
\begin{small}
\begin{tabular}{lrrrrr}
& \texttt{\textbf{S:NS}} & \texttt{\textbf{\#AActs}} & \texttt{\textbf{\#SActs}} & \texttt{\textbf{\#Turns}} & \texttt{\textbf{\#Tokens}}\\
\toprule
Few-Shot & 1 : \textcolor{red}{\textbf{0.4}} & 56 & 39 & 47 & \textcolor{red}{\textbf{856}}\\
\hdashline
& 1 : 2.3 & 89 & 27 & 48 & 1,609\\
\textsc{SportsGen} & 1 : \textcolor{blue}{\textbf{3.1}} & 97 & 26 & 45 & \textcolor{blue}{\textbf{1,700}}\\
(Ours) & 1 : 4.0 & 103 & 20 & 34 & 1,757\\
& 1 : 5.2 & 102 & 16 & 27 & 1,722\\
\hdashline
Human Narr. & 1 : \textcolor{blue}{\textbf{3.0}} & 114 & 29 & 65 & \textcolor{blue}{\textbf{1,632}}\\
\bottomrule
\end{tabular}
\end{small}
\vspace{-0.05in}
\caption{We compare human narratives with those of SportsGen and few-shot prompting. Prompting generates narratives deviating from typical basketball games. 
}
\label{tab:human_synthetic}
\vspace{-0.2in}
\end{table}

In Table~\ref{tab:results_real}, we report accuracy and DCA scores for \textsc{SportsGen} narratives with a 1:3 S:NS ratio, utilizing three strategies: DnC-10 (dividing the narrative into batches of 10 actions), DnC-P (division by individual players), and monolithic processing (Mono). We observe that accuracy for synthetic narratives tends to be lower than for NBA narratives. This may be due to the models' greater familiarity with NBA narratives, which may be included in the pretraining data, compared to our novel synthetic narratives. Further, our use of GPT-4o enhanced action templates introduces more lexical variety than the typical NBA narrative, creating a more challenging task for the models in accurately tracking team scores. Overall, Claude-3-Opus achieves the highest performance across all settings when measured by DCA scores, suggesting its effectiveness where some margin of error is permissible.


\vspace{0.05in}
\noindent\textbf{Qualitative Evaluation.}\,\, We conduct a qualitative comparison between \textsc{SportsGen} and real NBA narratives. A human expert or GPT-4 review 100 narrative snippets from each source, with each snippet containing 10 plays. The evaluation focuses on (a) the naturalness of the language and (b) the logical coherence of the narratives. Evaluators choose their preferred narrative or opt for a tie, also providing justifications. We shuffle the order of pairs to reduce bias in this process, i.e., it is not guaranteed that Snippet A $\succ$ Snippet B when (A, B) is presented to the human or system. The results show that our human expert prefers SportsGen narratives 39\% of the time, tied 9\%, and favors NBA narratives 52\% of the time. GPT-4's analysis was quite similar, with 39\% preference for SportsGen, 7\% ties, and 54\% for NBA narratives. These insights suggest that while SportsGen is aligning with human narrative standards, some gaps remain.

\begin{figure}[t]
\centering
\includegraphics[width=2.8in]{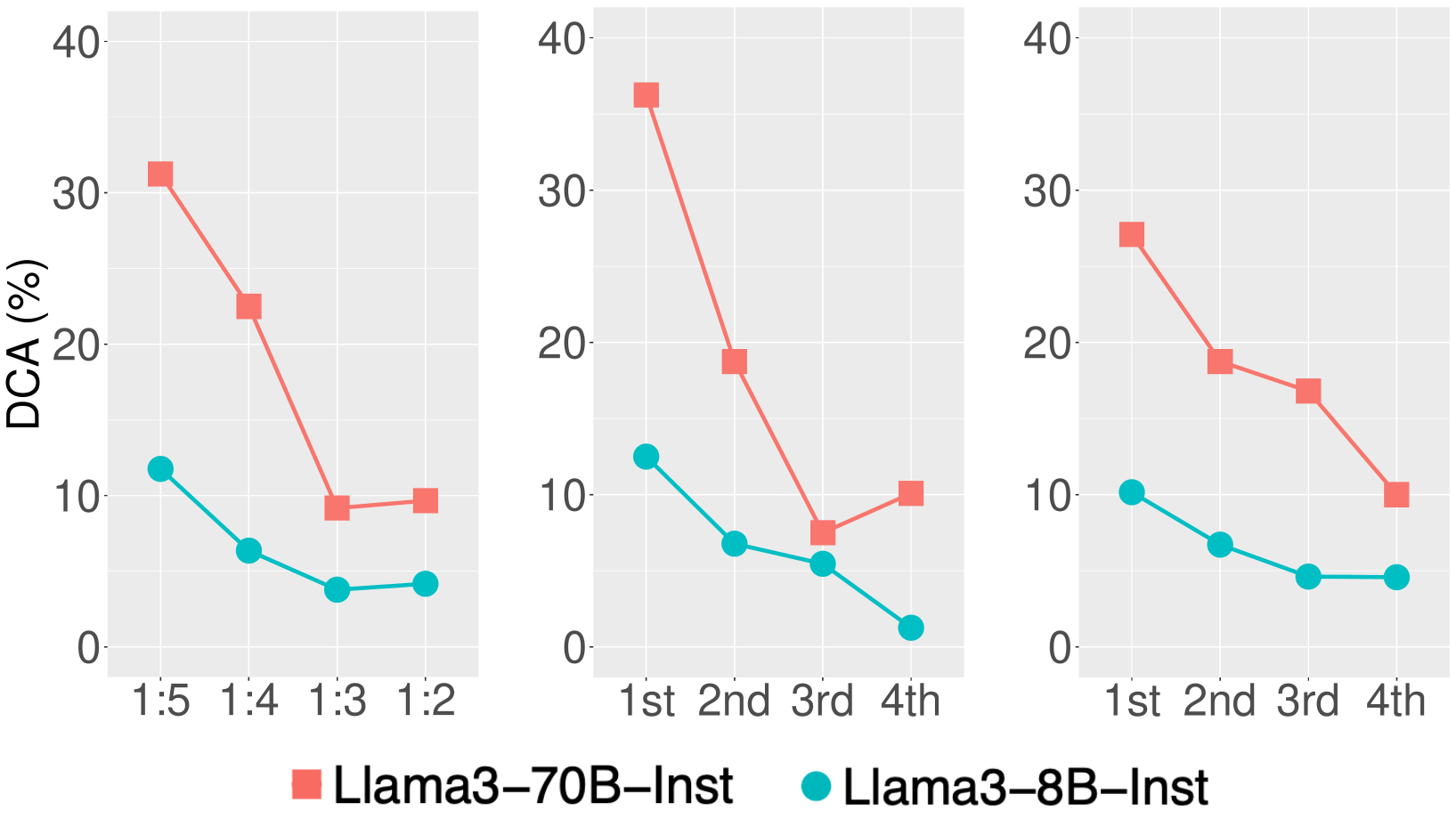}
\vspace{-0.05in}
\caption{(Left) Synthesized narratives with varying S:NS ratios. (Middle) Narratives grouped by \# of scoring actions. (Right) Narratives grouped by \# of tokens.
}
\vspace{-0.1in}
\label{fig:density}
\end{figure}

\subsection{Information Density on Reasoning Performance}
\label{sec:density}

Frequent scoring presents a challenge for LLMs in tracking and aggregating team points, especially during intense periods, e.g., the last quarter of a basketball game, where scoring spikes. In this study, we examine how models handle game quarters with varying levels of information density. We apply three settings: (a) four synthesized narrative sets with S:NS ratios of 1:5, 1:4, 1:3, and 1:2, where 1:2 represents a higher proportion of relevant information (see Figure~\ref{fig:density}, left); (b) we group synthesized narratives into four quartiles based on the total number of scoring actions (middle); and (c) for comparison, we also group narratives into four quartiles based on their length, measured in token count (see Figure~\ref{fig:density}, right).

In Figure~\ref{fig:density}, we report DCA scores for Llama3-8B-Inst and Llama3-70B-Inst to see how LLMs perform across various information densities. We note a decrease in performance as the models process inputs from low to high levels of information density, particularly in Llama3-70B-Inst, which exhibits the most noticeable decline at high S:NS ratios, such as 1:2 or 1:3. When analyzing narratives by length, the models exhibit a more gradual performance decrease. Our analysis indicates that \emph{a model's analytical reasoning capabilities are linked to the quantity and density of relevant facts}, echoing Greg Kamradt's needle-in-a-haystack test, where the challenge is to retrieve a single fact from an LLM's extensive context window.\footnote{\footnotesize \url{github.co/gkamradt/LLMTest_NeedleInAHaystack}}

\begin{figure}[h]
\centering
\includegraphics[width=1\linewidth]{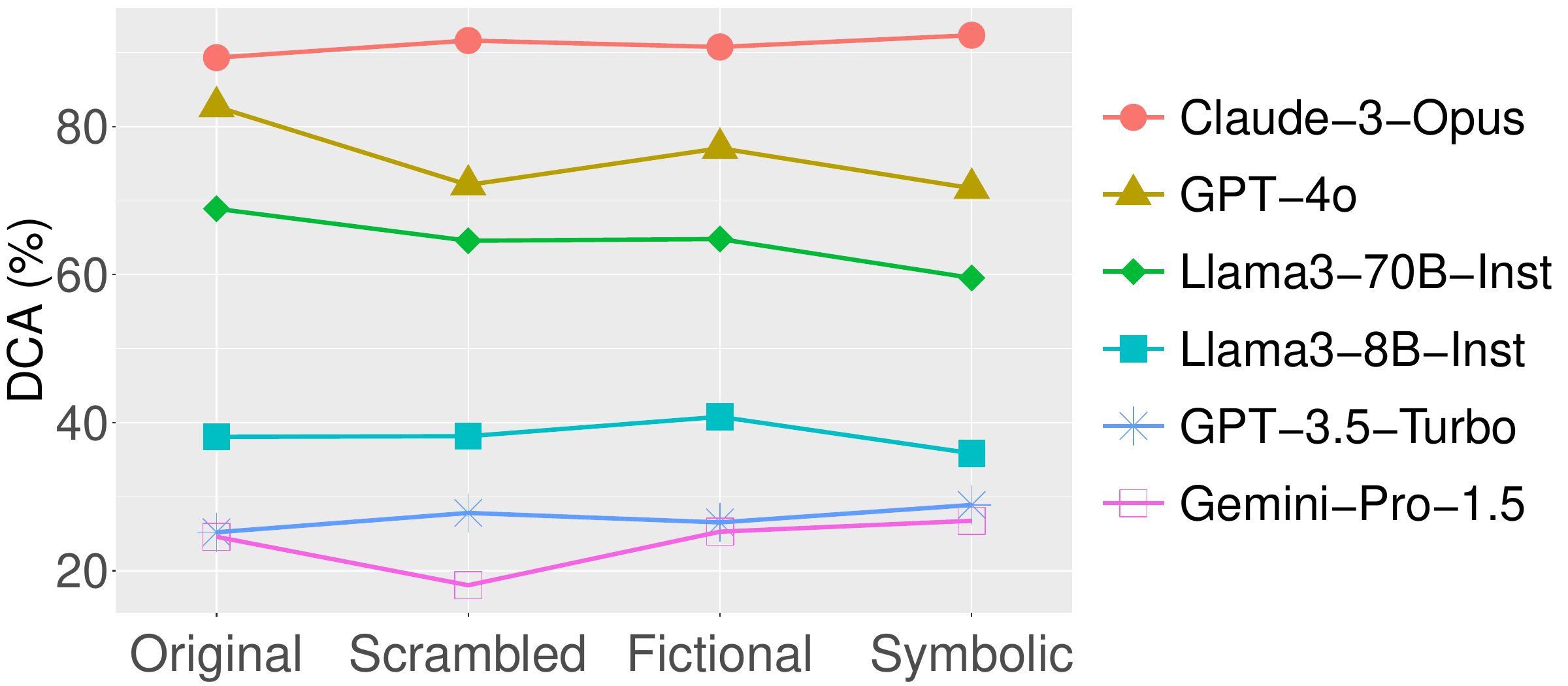}
\caption{We examine four scenarios that increasingly alter the context to a more `symbolic' format and evaluate the models' ability to calculate team points.}
\label{fig:symbolic}
\end{figure}

\subsection{From Natural Language to Symbolic Reasoning}
\label{sec:symbolic}

Symbolic reasoning uses symbols and logic to form traceable reasoning steps. Research has explored whether LLMs are capable of symbolic reasoning, or if they only infer patterns from data without explicit symbolic manipulation~\cite{gaur-saunshi-2023-reasoning,pan-etal-2023-logic,do-etal-2024-constraintchecker}. In this study, we examine four scenarios that increasingly alter the context to convert the original narratives to a more `symbolic' format. `\emph{Original}' are synthetic narratives using NBA team profiles; `\emph{Scrambled}' mismatch NBA players and teams; `\emph{Fictional}' replaces NBA player names with imaginary ones from FIFA; and `\emph{Symbolic}' substitutes players with labels like `Player 1, 2, 3…'. These changes are not expected to impact the model's ability to understand scoring actions or attributing them to teams. The actions described in play-by-plays remain unchanged as they characterize relationships between symbols, such as `Player 1 assists Player 2.' This method makes the game input less reliant on natural language and more on symbolic elements.

We examine the models' ability to calculate team points using synthetic narratives and present the DCA scores in Figure~\ref{fig:symbolic}. Our findings indicate that replacing original players with fictional names or indices can variably impact model performance. Models such as Llama3-8b-Inst and Gemini-Pro-1.5 score too low for us to draw definitive conclusions. Even high-performing models such as GPT-4o and Llama3-70b-Inst experience some decline in performance. Claude-3-Opus has demonstrated greater resilience than GPT-4o when faced with altered contexts. These results suggest that while LLMs have substantial reasoning skills, they rely on natural language context to perform well; thus, symbolic reasoning has room for improvement.

\section{Conclusion}
\label{sec: conclusion}

We investigate LLMs' analytical reasoning abilities, using divide-and-conquer strategies to determine where they perform best in analyzing sports narratives. With \textsc{SportsGen}, we further generate a variety of narratives that vary in style, complexity, and detail to better assess LLM performance. Our research not only paves the way for future LLM assessments but also sets an understanding of how LLMs may handle complex, longitudinal data.

\section{Limitations}

This study offers important insights into how LLMs handle sports narratives and there are a few limitations to consider. The use of synthesized narratives from \textsc{SportsGen}, while valuable for controlled experimentation, may not capture all the nuances of natural game commentary. Differences between our synthetic data and real-world narratives could influence how well findings generalize to natural settings. Additionally, our analysis is confined to basketball, a sport with frequent scoring events, which may not be representative of other sports that feature different dynamics and scoring patterns. Therefore, the broader applicability of our results to such diverse contexts is yet to be explored. Our future work may explore qualitative aspects such as strategic analysis or player performance that go beyond numerical scores. While acknowledging these limitations, our study still enriches the discussion about applying advanced reasoning techniques to sports narratives and sets the stage for future research in this exciting area.

\bibliography{custom,reasoning}

\appendix

\section{Example Prompts}
\label{sec:example_prompts}
In this section, we present sample prompts used in our experiments. Particularly, few-shot prompting uses existing play-by-plays to guide GPT-4o in creating synthesized game quarters. This method tends to result in repetitive scoring actions.

\begin{figure*}
\centering
\includegraphics[width=6.2in]{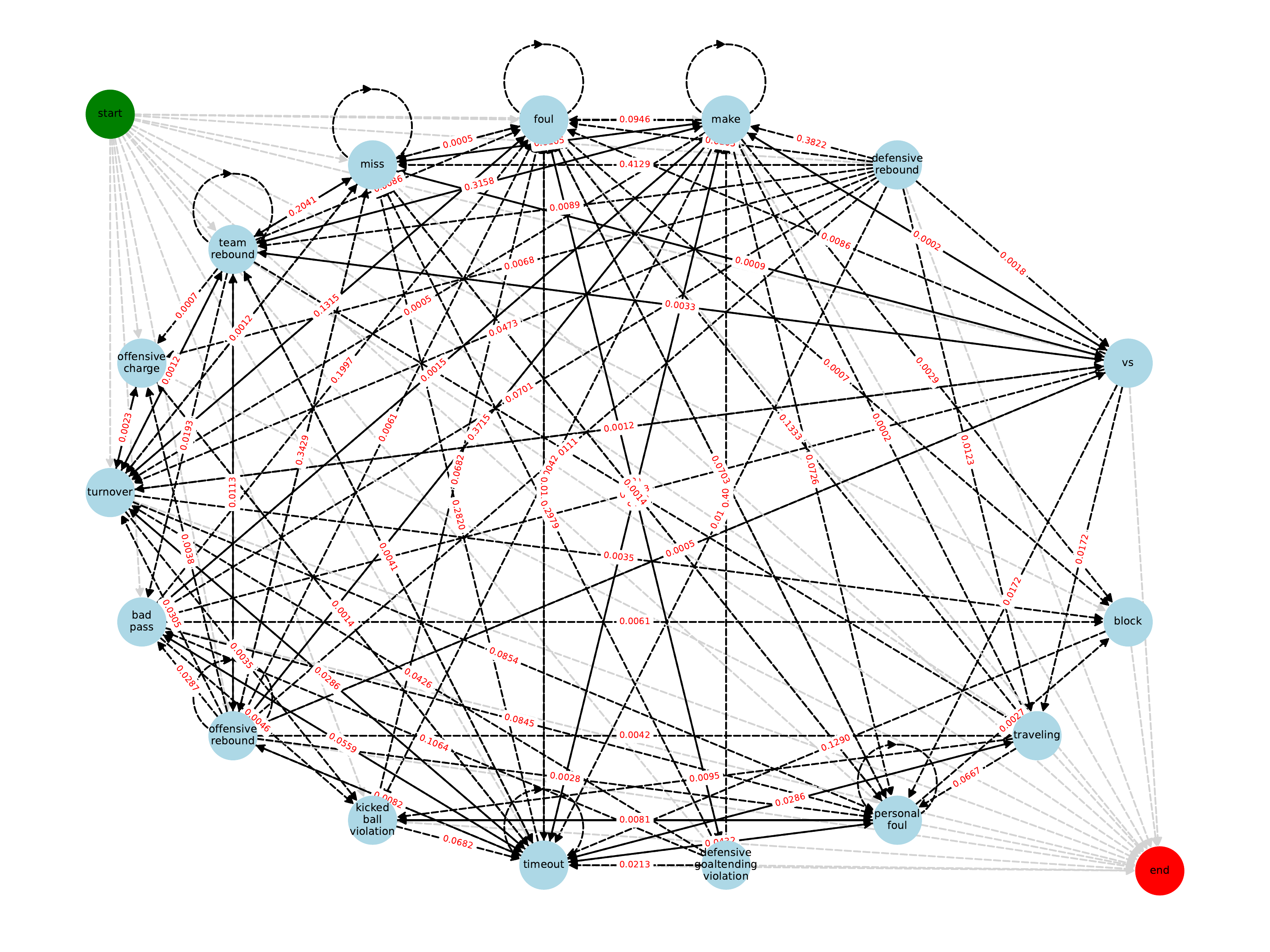}
\vspace{-0.4in}
\caption{An example action graph created from NBA narratives. When a team has the ball, they are on offense and use actions such as passing, dribbling, shooting to score points. The defense tries to stop them by blocking, stealing, and rebounding. We refer to each sequence of these actions as a \textbf{turn}, and characterize it using a Markov graph. In the graph, each significant action is represented as a node, and transitions between nodes show how team members cooperate and execute their tactics. The node `vs' denotes a matchup between two players, such as `Nikola Jokic vs. Anthony Davis,' indicating their direct competition in key moments of the game. The graph begins and ends with special nodes that mark the start and end of a turn.}
\vspace{-0.1in}
\label{fig:action_graph}
\end{figure*}

\begin{figure*}
\small
    \begin{samplebox}[label=pbp_1:3]{\texttt{SportsGen Narratives (S:NS Ratio = 1:3)}}
    \texttt{<Example 1>}\\
    \texttt{3:20    \quad Zach LaVine secured the defensive rebound.}\\
    \texttt{3:20    \quad Nikola Vucevic performs a dunk with an assist from Lonzo Ball.}\\
    \texttt{2:59    \quad LaMelo Ball fails to make a 17-foot jump shot.}\\
    \texttt{2:56    \quad Cody Martin secures an offensive rebound.}\\
    \texttt{2:38    \quad Mark Williams shoots and scores a flagrant 1 free throw.}\\
    \texttt{2:34    \quad Mark Williams committed a personal foul.}\\
    \texttt{2:33    \quad Lonzo Ball secures a defensive rebound.}\\
    \texttt{2:19    \quad DeMar DeRozan misses a 20-foot jumper.}\\
    \texttt{2:17    \quad Miles Bridges secures a defensive rebound.}\\
    \texttt{2:14    \quad Mark Williams fails to make a 17-foot pullup jump shot.}
      \\[1em]
    \texttt{<Example 2>}\\
    \texttt{3:00    \quad Jonathan Kuminga executes a tip shot.}\\
    \texttt{2:57    \quad Franz Wagner secured the defensive rebound.}\\
    \texttt{2:46    \quad Cole Anthony hit a 26-foot three pointer.}\\
    \texttt{2:39    \quad Stephen Curry fails to make a 29-foot three-point shot.}\\
    \texttt{2:38    \quad Golden State Warriors offensive team rebounded the ball.}\\
    \texttt{2:37    \quad Franz Wagner secured the defensive rebound.}\\
    \texttt{2:22    \quad Cole Anthony fails to make a 10-foot two-point shot.}\\
    \texttt{2:19    \quad Stephen Curry secures the defensive rebound.}\\
    \texttt{2:00    \quad Stephen Curry fails to connect on a 27-foot three-point shot.}\\
    \texttt{1:57    \quad Wendell Carter Jr. fails to make a 17-foot pullup jump shot.}\\  
    \end{samplebox}
\vspace{-0.1in}
\caption{\textsc{SportsGen} offers a practical option for creating realistic sports narratives.}
\end{figure*}

\begin{figure*}
\small
    \begin{minipage}{\textwidth}
    \begin{samplebox}[label=oneshot]{\texttt{Few-Shot Prompting Narratives}}
    \texttt{<Example 1>}\\
    \texttt{2:30    \quad Clint Capela makes 2-foot layup (Trae Young assists)}\\
    \texttt{2:15    \quad Jaylen Brown makes 3-point jumper (Derrick White assists)}\\
    \texttt{2:00    \quad Dejounte Murray makes 20-foot jumper}\\
    \texttt{1:45    \quad Jayson Tatum makes 18-foot jumper}\\
    \texttt{1:30    \quad Trae Young makes 3-point jumper}\\
    \texttt{1:15    \quad Kristaps Porzingis makes 10-foot jumper}\\
    \texttt{1:00    \quad Dejounte Murray makes 18-foot jumper}\\
    \texttt{0:45    \quad Jaylen Brown makes 2-foot layup (Jayson Tatum assists)}\\
    \texttt{0:30    \quad Trae Young makes 3-point jumper}\\
    \texttt{0:15    \quad Jayson Tatum makes 20-foot jumper}
     \\[1em]
    \texttt{<Example 2>}\\
    \texttt{8:13    \quad Rui Hachimura defensive rebound}\\
    \texttt{8:00    \quad D'Angelo Russell makes 20-foot jumper}\\
    \texttt{7:45    \quad Devin Vassell makes 25-foot three point jumper (Tre Jones assists)}\\
    \texttt{7:30    \quad LeBron James makes driving layup}\\
    \texttt{7:15    \quad Keldon Johnson makes 10-foot jumper}\\
    \texttt{7:00    \quad Anthony Davis makes 15-foot jumper}\\
    \texttt{6:45    \quad Victor Wembanyama makes 12-foot jumper}\\
    \texttt{6:30    \quad Austin Reaves makes 25-foot three point jumper (D'Angelo Russell assists)}\\
    \texttt{6:15    \quad Devin Vassell misses 20-foot jumper}\\
    \texttt{6:13    \quad Anthony Davis defensive rebound}\\  
    \end{samplebox}
    \end{minipage}
\caption{Few-shot prompting uses existing play-by-plays to guide GPT-4o in creating synthesized game quarter. This method tends to result in repetitive scoring actions.}

\end{figure*}

\begin{figure*}
\small
    \begin{minipage}{\textwidth}
    \begin{samplebox}[label=Success_example]{\texttt{An Example Showing a Successful Reasoning Attempt by GPT-4o}}
    \texttt{To determine the total points scored by each team, we need to examine the play-by-play descriptions and identify which team scored the points in each play. Here's the breakdown:}\\[1em]
    
    \texttt{1. {Jakob Poeltl makes layup (Keldon Johnson assists)} - San Antonio Spurs score 2 points.}\\
    
    \texttt{2. {Gordon Hayward makes 15-foot pullup jump shot} - Charlotte Hornets score 2 points.}\\
    
    \texttt{3. {Gordon Hayward makes free throw 1 of 1} - Charlotte Hornets score 1 point.}\\
    
    \texttt{4. {Terry Rozier makes 25-foot three point jumper (Mason Plumlee assists)} - Charlotte Hornets score 3 points.}\\
    
    \texttt{5. {Jalen McDaniels makes layup (Terry Rozier assists)} - Charlotte Hornets score 2 points.}\\
    
    \texttt{6. {Devin Vassell makes free throw 1 of 2} - San Antonio Spurs score 1 point.}\\
    
    \texttt{7. {Devin Vassell makes free throw 2 of 2} - San Antonio Spurs score 1 point.}\\
    
    \texttt{8. {Mason Plumlee makes alley oop dunk shot (James Bouknight assists)} - Charlotte Hornets score 2 points.}\\
    
    \texttt{9. {Gordon Hayward makes free throw 1 of 2} - Charlotte Hornets score 1 point.}\\
    
    \texttt{10. {Gordon Hayward makes free throw 2 of 2} - Charlotte Hornets score 1 point.}\\[1em]
    
    \texttt{Now, let's sum up the points for each team:}\\[1em]
    
    \texttt{- {Charlotte Hornets}:}\\
    \texttt{    2 (Hayward) + 1 (Hayward FT) + 3 (Rozier) + 2 (McDaniels) + 2 (Plumlee) + 1 (Hayward FT) + 1 (Hayward FT) = 12 points}\\[1em]
    
    \texttt{- {San Antonio Spurs}:}\\
    \texttt{    2 (Poeltl) + 1 (Vassell FT) + 1 (Vassell FT) = 4 points}\\[1em]
    
    \texttt{Here is the final result in JSON format:}\\
    \begin{verbatim}
    {
      "Charlotte Hornets": 12,
      "San Antonio Spurs": 4
    }
    \end{verbatim}

    \end{samplebox}
    \end{minipage}
\caption{An example showing a successful reasoning attempt by GPT-4o.}
\end{figure*}

\begin{figure*}
\small
    \begin{minipage}{\textwidth}
    \begin{samplebox}[label=fail_example]{\texttt{An Example Showing an Unsuccessful Reasoning Attempt by LLama-3-8B-Instruct}}
    \texttt{To determine the total points scored by each team, I'll analyze the play-by-play descriptions and identify the scoring plays.}\\[0.5em]
    
    \texttt{Charlotte Hornets:}
    
    \texttt{1. Gordon Hayward makes 15-foot pullup jump shot (5:53) - 2 points}
    
    \texttt{2. Terry Rozier makes 25-foot three point jumper (5:32) - 3 points}
    
    \texttt{3. Jalen McDaniels makes layup (5:06) - 2 points}
    
    \texttt{4. Mason Plumlee makes alley oop dunk shot (4:43) - 2 points}
    
    \texttt{5. Gordon Hayward makes free throw 1 of 2 (4:24) - 1 point}
    
    \texttt{6. Gordon Hayward makes free throw 2 of 2 (4:24) - 1 point}
    
    \texttt{Total points for Charlotte Hornets: 2 + 3 + 2 + 2 + 1 + 1 = 11 points}\\[0.5em]
    
    \texttt{San Antonio Spurs:}
    
    \texttt{1. Jakob Poeltl makes layup (6:10) - 2 points}
    
    \texttt{2. Devin Vassell makes free throw 1 of 2 (4:55) - 1 point}
    
    \texttt{3. Devin Vassell makes free throw 2 of 2 (4:55) - 1 point}
    
    \texttt{Total points for San Antonio Spurs: 2 + 1 + 1 = 4 points}\\[0.5em]
    
    \texttt{Here is the final result in JSON format:}
    
   \begin{verbatim}
    {
      "Charlotte Hornets": 11,
      "San Antonio Spurs": 4
    }
    \end{verbatim}
    \end{samplebox}
    \end{minipage}
\vspace{-0.05in}
\caption{An example showing an unsuccessful reasoning attempt by LLama-3-8B-Instruct.}
\end{figure*}

\begin{figure*}[htbp]
\centering
\begin{footnotesize}
\begin{minipage}{\textwidth}

\begin{promptbox}[label=system]{\texttt{System Message}}
\texttt{You are a helpful assistant tasked with analyzing sports games. You have been given a play-by-play breakdown of an NBA basketball game between two teams.}\\[0.5em]
\texttt{The ``Time'' column shows the exact time on the game clock when each play took place. The game clock counts down, so this column displays times in a descending order.}\\[0.5em]
\texttt{The ``Play'' column describes the action that happened at the respective times. It provides details of specific plays, movements, and outcomes on the court.}\\[0.5em]
\texttt{Team players are listed in two rows, each row representing one of the two basketball teams involved in the game.}
\end{promptbox}

\begin{promptbox}[label=user]{\texttt{User Message}}
\texttt{Analyze the team-player affiliations and play-by-play descriptions below to determine the total points scored by each team.}\\[0.5em]
\texttt{Please explain your reasoning step by step and provide the final results in the JSON format. Starting with: \{Toronto Raptors: 0, Indiana Pacers: 0\}}\\[1em]
\texttt{\#Team-Player Affiliations:}\\[0.5em]
\texttt{Toronto Raptors: Og Anunoby, Fred Vanvleet, Norman Powell, Aron Baynes, Chris Boucher, Deandre Bembry, Stanley Johnson, Yuta Watanabe, Malachi Flynn, Matt Thomas, Terence Davis}\\[0.5em]
\texttt{Indiana Pacers: Myles Turner, Justin Holiday, Doug Mcdermott, Jeremy Lamb, Malcolm Brogdon, Domantas Sabonis, TJ McConnell, Aaron Holiday}\\[1em]
\texttt{\#Play-by-Play Descriptions:}\\ [0.5em]
\texttt{Time    \,\,\,\quad Play }\\
\texttt{12:00   \quad Aron Baynes vs. Myles Turner (Fred VanVleet gains possession)}\\
\texttt{11:38   \quad Myles Turner blocks Stanley Johnson's two point shot}\\
\texttt{11:35   \quad Justin Holiday defensive rebound}\\
\texttt{11:24   \quad Fred VanVleet blocks Malcolm Brogdon's two point shot}\\
\texttt{11:22   \quad Fred VanVleet defensive rebound}\\  
\texttt{11:20   \quad Aron Baynes makes layup (Fred VanVleet assists)}\\
\texttt{11:04   \quad Stanley Johnson shooting foul}\\
\texttt{11:04   \quad Myles Turner misses free throw 1 of 2}\\
\texttt{11:04   \quad Pacers offensive team rebound}
\end{promptbox}
\end{minipage}
\end{footnotesize}
\vspace{-0.1in}
\caption{Our research uses analytical reasoning to accurately calculate team points from sports narratives.}
\end{figure*}

\begin{figure*}[htbp]
    \centering
    \begin{footnotesize}

    \begin{minipage}{\textwidth}
    \begin{promptbox}[label=system]{\texttt{System Message}}
    \texttt{Your task is to generate detailed play-by-play descriptions for one quarter of an NBA game, capturing the dynamic flow of the game. This includes player actions (e.g., shooting, blocking, stealing, etc.), key moments (e.g., turnovers, fouls, timeouts), player substitutions, and more. Please ensure the narrative is engaging and reflects the pace and intensity of an NBA game.}
    \end{promptbox}
    \begin{promptbox}[label=user]{\texttt{User Message}}
    \texttt{Generate a full quarter of an NBA game using the detailed player-team affiliations and play-by-play data provided. Ensure the following requirements are met:}\\[1em]
    \texttt{1. Typically, a quarter of NBA games feature 70 to 150 plays, with 20 to 40 being scoring}\\
    \quad\,\,\,\texttt{plays, and a scoring density between 0.25 and 0.35.}\\
    \texttt{2. A team's efficiency score, showing potential points per 100 possessions,}\\
    \quad\,\,\,\texttt{can impact game outcomes.}\\
    \texttt{3. Adhere to the example play-by-play description format provided.}\\
    \texttt{4. Ensure a natural and logical progression between plays.}\\
    \texttt{5. Separate each column in the play-by-play data using a tab character.}\\[1em]
    \texttt{\#Team-Player Affiliations:} \\[0.5em]
    \texttt{Team-1: player-1, player-2, player-3, player-4, player-5, player-6, player-7, ...}\\
    \texttt{Team-2: player-11, player-12, player-13, player-14, player-15, player-16, player-17, ...}
    \\[1em]
    \texttt{\#Team Efficiency Scores:}\\[0.5em]
    \texttt{Team-1: 93}\\
    \texttt{Team-2: 87}\\[1em]
    \texttt{\#Example Play-by-Play Descriptions:}\\[0.5em]
    \texttt{Time    \,\,\,\quad Team    \,\,\,\,\,\,\quad Play}\\
    \texttt{12:00    \quad Team-2    \quad player-3 vs. player-14 (player-13 gains possession)}\\
    \texttt{11:46    \quad Team-2    \quad player-12 makes alley oop layup (player-14 assists)}\\
    \texttt{11:26    \quad Team-1    \quad player-2 makes 2-foot layup (player-3 assists)}\\
    \texttt{11:08    \quad Team-2    \quad player-12 makes driving layup}\\
    \texttt{10:43    \quad Team-1    \quad player-6 misses 21-foot step back jumpshot}\\
    \texttt{10:41    \quad Team-1    \quad player-2 offensive rebound}\\
    \texttt{10:37    \quad Team-2    \quad player-13 misses 25-foot three point jumper}\\
    \texttt{10:34    \quad Team-2    \quad player-15 defensive rebound}\\
    \texttt{10:27    \quad Team-2    \quad player-14 shooting foul}\\
    \texttt{10:27    \quad Team-2    \quad player-7 misses free throw 1 of 2}\\
    \texttt{10:27    \quad Team-1    \quad team-1 offensive team rebound}\\
    \texttt{10:27    \quad Team-2    \quad player-12 makes free throw 2 of 2}\\
    \texttt{10:14    \quad Team-2    \quad player-18 misses 23-foot three point jumper}\\
    \texttt{10:11    \quad Team-1    \quad player-6 defensive rebound}\\
    ...\\[1em]
    \texttt{Based on the following team-player affiliations, generate the play-by-plays for the first quarter of the game.}\\[1em]
    \texttt{\#Team-Player Affiliations:}\\[0.5em]
    \texttt{Memphis Grizzlies: Jordan Goodwin, Scotty Pippen Jr., Jake LaRavia,}\\ \quad\quad\quad\quad\quad\quad\quad\quad\quad\,\,\,\texttt{Vince Williams Jr., Santi Aldama}\\
    \texttt{Miami Heat: Bam Adebayo, Jimmy Butler, Jamal Cain, Haywood Highsmith, Jaime Jaques Jr.}\\[1em]
    \texttt{\#Team Efficiency Scores:}\\[0.5em]
    \texttt{Memphis Grizzlies: 92}\\
    \texttt{Miami Heat: 87}\\[1em]
    \texttt{\#Play-by-Play Descriptions:}\\[0.5em]
    \end{promptbox}
    \end{minipage}
    \end{footnotesize}
\caption{Few-shot prompting uses existing play-by-plays to guide GPT-4o in creating synthesized game quarters.}
\end{figure*}

\end{document}